\documentclass[lettersize,journal]{IEEEtran}
\usepackage{amsmath,amsfonts}
\usepackage{algorithmic}
\usepackage{algorithm}
\usepackage{array}
\usepackage[caption=false,font=normalsize,labelfont=sf,textfont=sf]{subfig}
\usepackage{textcomp}
\usepackage{stfloats}
\usepackage{url}
\usepackage{verbatim}
\usepackage{graphicx}
\usepackage{cite}
\usepackage{xcolor}
\usepackage{multirow}
\usepackage{booktabs} 
\usepackage[colorlinks=true, linkcolor=blue, citecolor=blue, urlcolor=blue]{hyperref}
\makeatletter
\def\@cite#1#2{{\hypersetup{citecolor=blue}\textcolor{blue}{[{#1\if@tempswa , #2\fi}]}}}
\makeatother
\usepackage{amssymb}
\usepackage[table]{xcolor}  
\usepackage{colortbl}       
\begin{document}

\title{MDAFNet: Multiscale Differential Edge and Adaptive Frequency Guided Network for Infrared Small Target Detection}

\author{Shuying~Li, Qiang~Ma, San~Zhang, Wuwei~Wang, Chuang~Yang,~\IEEEmembership{Member,~IEEE}
	\thanks{Shuying Li, Qiang Ma, San Zhang and Wuwei Wang are with	the School of Artificial Intelligence and School of Automation, Xi’an University of Posts and Telecommunications, Xi’an 710121, China, and also with Shanghai Artificial Intelligence Laboratory, Shanghai 200232, China(e-mail: lishuying@xupt.edu.cn; maqiang123@stu.xupt.edu.cn;	zhangsan@xupt.edu.cn; wangwuwei@xupt.edu.cn). Chuang Yang is with the Department of Electrical and Electronic Engineering, The Hong Kong	Polytechnic University, Hong Kong SAR (e-mail: omtcyang@gmail.com).}%
	\thanks{Corresponding author: Chuang Yang.}
}

\markboth{~}%
{Shell \MakeLowercase{\textit{et al.}}: A Sample Article Using IEEEtran.cls for IEEE Journals}


\maketitle

\begin{abstract}
Infrared small target detection (IRSTD) plays a crucial role in numerous military and civilian applications. However, existing methods often face the gradual degradation of target edge pixels as the number of network layers increases, and traditional convolution struggles to differentiate between frequency components during feature extraction, leading to low-frequency backgrounds interfering with high-frequency targets and high-frequency noise triggering false detections. To address these limitations, we propose MDAFNet (Multi-scale Differential Edge and Adaptive Frequency Guided Network for Infrared Small Target Detection), which integrates the Multi-Scale Differential Edge (MSDE) module and Dual-Domain Adaptive Feature Enhancement (DAFE) module. The MSDE module, through a multi-scale edge extraction and enhancement mechanism, effectively compensates for the cumulative loss of target edge information during downsampling. The DAFE module combines frequency domain processing mechanisms with simulated frequency decomposition and fusion mechanisms in the spatial domain to effectively improve the network's capability to adaptively enhance high-frequency targets and selectively suppress high-frequency noise. Experimental results on multiple datasets demonstrate the superior detection performance of MDAFNet.
\end{abstract}


\section{Introduction}
\IEEEPARstart{I}{RSTD} is a critical technology in remote sensing and defense, playing an essential role in early warning systems, search and rescue operations, and security surveillance. Infrared imaging sensors can achieve detection under complex conditions such as low visibility, nighttime, and adverse weather by capturing the thermal radiation of targets.However, due to the inherent characteristics of infrared small targets~\cite{9714770}, these targets appear as indistinct pixels with small information proportions and minimal texture features. The challenge of accurate detection and positioning is further compounded when they are masked by complex background clutter.

In light of these challenges, researchers have developed a range of traditional IRSTD methods, which are generally categorized into filtering and background suppression strategies~\cite{deshpande1999max}, human visual system (HVS)-inspired approaches~\cite{6479296}, and low-rank sparse representation techniques for background modeling~\cite{6595533}. However, the limited feature representation capability and reliance on manually designed priors result in poor performance when dealing with complex scenes. 

In contrast, deep learning approaches have demonstrated superior performance in IRSTD by automatically learning discriminative features from data. Among deep learning-based methods, Convolutional Neural Network (CNN)-based methods~\cite{9423171,9864119,9989433,10658560,yang2025pinwheel,10011452,10955237,11106397,yang2023instance} employ encoder-decoder architectures. For example, ACM-Net~\cite{9423171} promoted multi-scale feature interaction across hierarchical layers, while UIU-Net~\cite{9989433} employed a hierarchical U-shaped architecture for dense cross-level feature interaction.  Transformer methods~\cite{zhang2022rkformer,10924678,10219645,10011449,10486932} employ self-attention for global context modeling. RKformer~\cite{zhang2022rkformer} combined Transformer and convolution in parallel, leveraging the Runge-Kutta method for feature enhancement. STPSA-Net~\cite{10924678} leveraged semantic tokens and patch-wise spatial attention for accurate localization. However, most models undergo multiple downsampling operations during detection. As network depth increases through repeated downsampling operations, the edge information of small targets undergoes cumulative degradation across network layers. This considerably affects subsequent feature extraction and target localization accuracy. Moreover, traditional convolutions naturally possess smoothing characteristics, lacking explicit frequency processing mechanisms, which causes background clutter and noise interference. In addition, recent frameworks~\cite{11017756,11183610} combined CNNs with frequency domain processing. HDNet~\cite{11017756} adopted a dynamic high-pass filter to progressively filter low-frequency background information in the frequency domain, alleviating background interference to some extent. FAA-Net~\cite{11183610} employed DCT to capture high-frequency components and enhance local contrast. However, existing frequency domain methods have several limitations. They only use conventional frequency transforms (e.g., Fourier or DCT) with fixed filtering strategies lacking layer-adaptive modulation. Moreover, these methods lack the ability to differentially model frequency components across directions, and struggle to simultaneously adapt to the distinct frequency requirements of shallow layers retaining high-frequency details and deep layers suppressing high-frequency noise while maintaining semantics. In the edge domain, ISNet~\cite{zhang2022isnet} designed ODE-inspired edge blocks for shape extraction, but lacks multi-scale differential enhancement capability.

To mitigate the limitations identified above, we develop MDAFNet, which contains two core modules. First, the Multi-Scale Differential Edge (MSDE) module constructs an independent edge branch that performs three-way fusion with main branch features at skip connections to offset the cumulative loss of target edge details in the downsampling process of the main branch. Second, the Dual-Domain Adaptive Feature Enhancement (DAFE) module achieves adaptive frequency-guided feature enhancement at skip connections, enabling the network to adaptively enhance high-frequency targets and selectively suppress high-frequency noise. The combination of both thereby enhances the network's detection performance for infrared small targets.

We summarize our key contributions below:

1) We design the MSDE module with multi-scale differential edge enhancement mechanisms to effectively maintain target geometric detail integrity.

2) We develop the DAFE module with adaptive frequency-guided enhancement to effectively discriminate targets from high-frequency noise through dual-domain processing.

3) Extensive experiments across various benchmark datasets show that MDAFNet outperforms state-of-the-art methods, confirming our approach's effectiveness.

\section{METHOD}
\subsection{Overall architecture}
\begin{figure*}[!t]
	\centering
	\includegraphics[width=\textwidth]{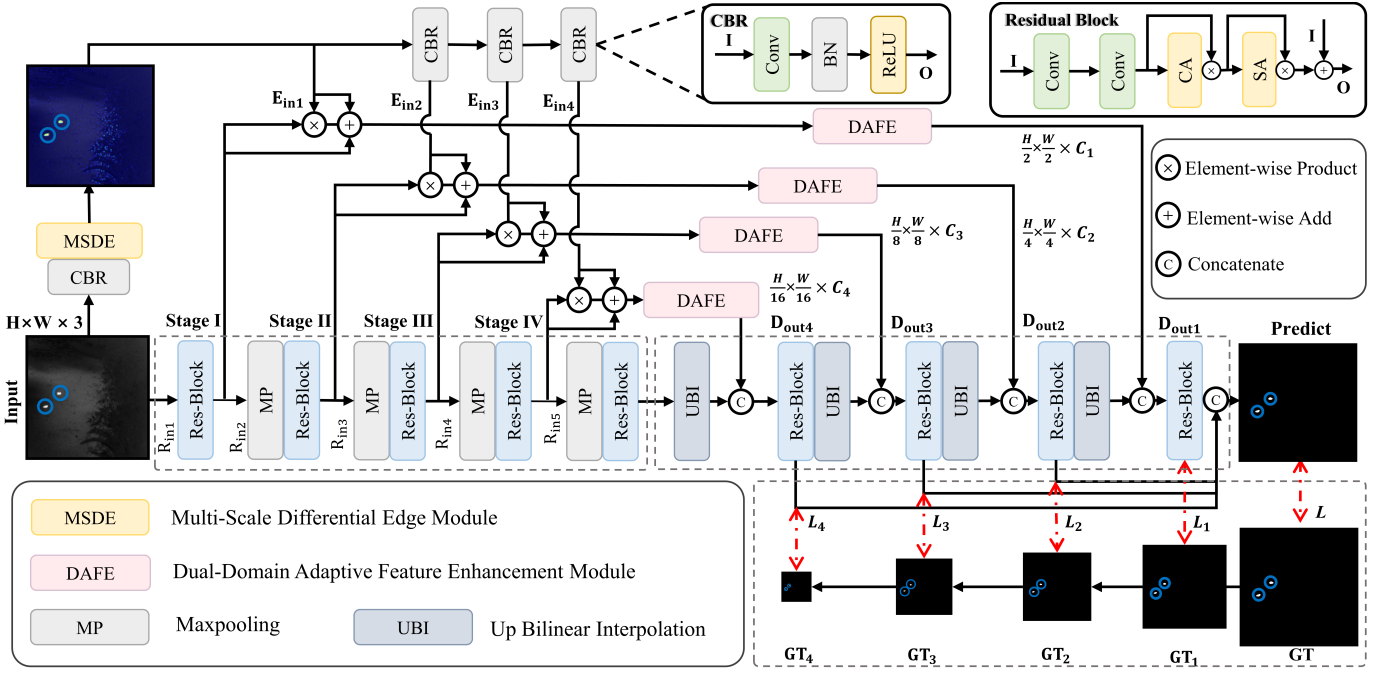}
	\caption{Schematic diagram of the MDAFNet framework.} 
	\label{Fig.1}
\end{figure*}
As shown in Fig.~\ref{Fig.1}, MDAFNet employs a U-shaped structure for IRSTD. The encoder contains four stages (Stage I-IV), where each stage employs Residual Blocks (Res-Block) for feature extraction, with feature map resolutions progressively decreasing. To address the cumulative loss of edge information during downsampling and the inadequacy of frequency processing, the network integrates two core modules: MSDE and DAFE. The MSDE module constructs an independent auxiliary edge branch that extracts multi-level edge features ($\mathbf{E}_{\mathrm{in}1}$ to $\mathbf{E}_{\mathrm{in}4}$) and performs adaptive triple-path fusion at skip connections using main branch features, edge features, and the multiplication of main branch and edge features, effectively compensating for edge information loss. The DAFE module is deployed at each skip connection to enhance the network's capability of adaptively enhancing high-frequency targets and selectively suppressing high-frequency noise through adaptive frequency-guided enhancement. The decoder progressively recovers spatial resolution and fuses multi-scale features through skip connections. The network employs a deep supervision strategy, utilizing SLS Loss~\cite{10658560} to supervise the outputs of each decoder stage, with the output of the final decoder layer serving as the prediction map.
\subsection{Multi-Scale Differential Edge Module}
\begin{figure}[!t]
	\centering
	\includegraphics[width=0.48\textwidth]{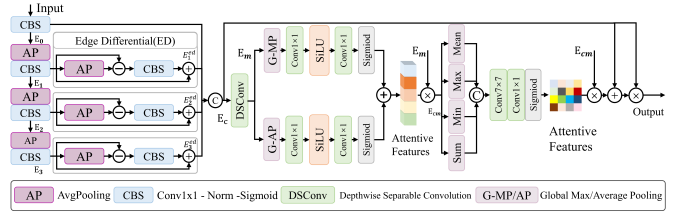}
	\caption{Architecture of the MSDE module.} 
	\label{Fig.2}
\end{figure}
To compensate for edge information loss during downsampling, we propose 
the MSDE. Specifically, given an input feature $\mathbf{X}$, MSDE first projects it into the edge feature space. Then, it employs hierarchical average pooling to extract edge features at multiple scales $(t = 1, 2, 3)$. Notably, at each scale, an Edge Differential (ED) strengthens edge perception through differential operations. Then, the initial feature and edge-enhanced features at all scales are concatenated and fused through a fusion module to generate a comprehensive edge representation $\mathbf{E}^{c}$. The operation is defined as:
\begin{gather}
	\mathbf{E}_0 = \mathrm{CBS}(\mathbf{X}),\\
	\mathbf{E}_t = \mathrm{AP}(\mathrm{CBS}(\mathbf{E}_{t-1})),\\
	\mathbf{E}^{\mathrm{ed}}_t = \mathbf{E}_t + \mathrm{CBS}(\mathbf{E}_t - \mathrm{AP}(\mathbf{E}_t)),\\
	\mathbf{E}^{c} = \mathrm{U}([\mathbf{E}_0, \mathbf{E}^{\mathrm{ed}}_1, \mathbf{E}^{\mathrm{ed}}_2, \mathbf{E}^{\mathrm{ed}}_3]),
\end{gather}
where $\mathrm{AP}$ denotes average pooling with kernel size $3 \times 3$ 
and stride $1$, $\mathrm{CBS}$ represents a $1 \times 1$ convolution with batch normalization followed by sigmoid, and $\mathrm{U}$ denotes channel concatenation followed by $1\times1$ convolution.

To enhance multi-scale edge features, we employ channel and spatial attention for refinement. The channel attention captures inter-channel dependencies through dual-path pooling, while the spatial attention leverages multiple statistics to emphasize salient regions. The operation is defined as:
\begin{gather}
	\mathbf{E}^{m} = \mathrm{DSConv}(\mathbf{E}^{c}), \\
	\mathbf{E}^{\mathrm{ca}}_{\mathrm{out}} = \mathbf{E}^{m} \odot \sigma(\phi(\mathrm{G\text{-}AP}(\mathbf{E}^{m})) + \phi(\mathrm{G\text{-}MP}(\mathbf{E}^{m}))), \\
	\mathbf{E}^{\mathrm{sa}}_{\mathrm{out}} = \mathbf{E}^{\mathrm{ca}}_{\mathrm{out}} \odot \sigma(\mathrm{Conv}_{1\times1}(\mathrm{SiLU}(\mathrm{Conv}_{7\times7}(\mathcal{M}(\mathbf{E}^{\mathrm{ca}}_{\mathrm{out}}))))),
\end{gather}
where $\mathrm{DSConv}$ is depthwise separable convolution, $\phi$ denotes two $1 \times 1$ convolutions with SiLU, $\sigma$ is sigmoid, and $\mathcal{M}$ concatenates mean, max, min, and sum pooling.

Finally, a residual connection with element-wise multiplication is applied. The operation is defined as:
\begin{equation}
	\mathbf{E}^{\mathrm{out}} = (\mathbf{E}^{\mathrm{sa}}_{\mathrm{out}} + \mathbf{E}^{c})\odot \mathbf{E}^{c}.
\end{equation}
\subsection{Dual-Domain Adaptive Feature Enhancement}
\begin{figure}[!t]
	\centering
	\includegraphics[width=0.48\textwidth]{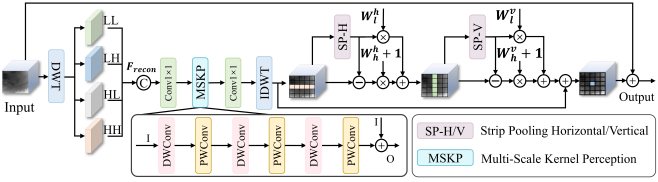}
	\caption{Architecture of the DAFE module.} 
	\label{Fig.3}
\end{figure}
To address the inadequacy of frequency processing in existing methods, we propose the DAFE module for adaptive frequency-guided 
feature enhancement. For a feature representation $\mathbf{F} \in \mathbb{R}^{C \times H \times W}$, DAFE first decomposes it into four frequency subbands using the Haar wavelet transform. These subbands are then concatenated, reduced in dimension, processed through the Multi-Scale Kernel Perception (MSKP) mechanism, and reconstructed to the spatial domain via inverse wavelet transform. The operation is defined as:
\begin{gather}
	\mathbf{F}_{\mathrm{ll}}, \mathbf{F}_{\mathrm{lh}}, \mathbf{F}_{\mathrm{hl}}, \mathbf{F}_{\mathrm{hh}} = \mathrm{DWT}(\mathbf{F}), \\
	\mathbf{F}_{\mathrm{mskp}} = \mathrm{MSKP}(\mathrm{Conv}_{1\times1}([\mathbf{F}_{\mathrm{ll}}, \mathbf{F}_{\mathrm{lh}}, \mathbf{F}_{\mathrm{hl}}, \mathbf{F}_{\mathrm{hh}}])), \\
	\mathbf{F}_{\mathrm{recon}} = \mathrm{IDWT}(\mathrm{Split}(\mathrm{Conv}_{1\times1}(\mathbf{F}_{\mathrm{mskp}}))),
\end{gather}
where $\mathbf{F}_{\mathrm{ll}} \in \mathbb{R}^{C \times \frac{H}{2} \times \frac{W}{2}}$ represents the low-frequency approximation, $\mathbf{F}_{\mathrm{lh}}$, $\mathbf{F}_{\mathrm{hl}}$, $\mathbf{F}_{\mathrm{hh}}$ represent horizontal, vertical, and diagonal high-frequency details, respectively, and MSKP employs stacked depthwise separable convolutions at three scales $(3, 5, 7)$ for multi-scale feature extraction across different receptive fields.

To further refine the frequency components, we apply an Adaptive Frequency Modulation (AFM) mechanism with two-stage strip pooling: horizontal (SP-H) and vertical (SP-V). In each stage, strip-shaped average pooling extracts low-frequency components, while high-frequency components are obtained by subtraction. These frequency components are then modulated using learnable channel-wise parameters. Taking the horizontal stage as an example:
\begin{gather}
	\mathbf{F}^{l}_{h} = \mathrm{SP\text{-}H}(\mathbf{F}_{\mathrm{recon}}),\quad \mathbf{F}^{h}_{h} = \mathbf{F}_{\mathrm{recon}} - \mathbf{F}^{l}_{h},\\
	\mathbf{F}_{h}^{\mathrm{out}} = \mathbf{w}^{h}_{l} \odot \mathbf{F}^{l}_{h} + (\mathbf{w}^{h}_{h} + 1) \odot \mathbf{F}^{h}_{h},
\end{gather}
where $\mathrm{SP\text{-}H}$ denotes horizontal strip pooling with kernel size $(1, K)$, $K$ is the strip length, and $\mathbf{w}^{h}_{l}, \mathbf{w}^{h}_{h} \in \mathbb{R}^{C}$ are learnable parameters for modulating low- and high-frequency components, respectively. The term $(\mathbf{w}^{h}_{h} + 1)$ ensures that high-frequency information is preserved. The vertical stage follows the same formulation with $\mathrm{SP\text{-}V}$ using kernel $(K, 1)$ and parameters $\mathbf{w}^{v}_{l}$, $\mathbf{w}^{v}_{h}$, yielding the final output $\mathbf{F}_{\mathrm{afm}}$.

Finally, the output features are integrated with the input through learnable weighted skip connections. The operation is defined as:
\begin{equation}
	\mathbf{F}_{\mathrm{out}} = \boldsymbol{\alpha} \odot \mathbf{F} + \boldsymbol{\beta} \odot \mathbf{F}_{\mathrm{afm}},
\end{equation}
where $\boldsymbol{\alpha}$ and $\boldsymbol{\beta} \in \mathbb{R}^{C}$ denote adaptive weights that balance the contributions of the original input and the modulated features.

\section{EXPERIMENTS}

\subsection{Datasets and Implementation Setup}

To thoroughly assess our method's performance, extensive evaluations are performed across three standard IRSTD benchmarks: IRSTD-1K, NUAA-SIRST, and SIRST-Aug. The network is trained using the Adagrad optimizer for 400 epochs with a learning rate initialized at 0.05 and batch size set to 4. For fair comparison, these hyperparameters remain consistent across all datasets. Experiments utilize an NVIDIA RTX 4090 GPU. Model performance is quantified using three metrics: $IoU$ (intersection over union), $P_d$ (probability of detection), and $F_a$ (false alarm rate).
\subsection{Comparsion With SOTA Methods}

For performance validation, our approach is compared against 11 SOTA methods, including 8 CNN-based approaches (DNA-Net~\cite{9864119}, UIU-Net~\cite{9989433}, RDIAN~\cite{10011452}, AGPC-Net~\cite{10024907}, MSHNet~\cite{10658560}, HCF-Net~\cite{10687431}, L2SKNet~\cite{10813615}, PConv~\cite{yang2025pinwheel}) and 3 Hybrid approaches (ABC~\cite{10219645}, MTU-Net~\cite{10011449}, SCTransNet~\cite{10486932}). Evaluations are performed across three benchmarks: IRSTD-1K, NUAA-SIRST, and SIRST-Aug, with results analyzed both quantitatively and qualitatively.

\subsubsection{Quantitative Evaluations Results}
\begin{figure}[!t]
	\centering
	\includegraphics[width=0.48\textwidth]{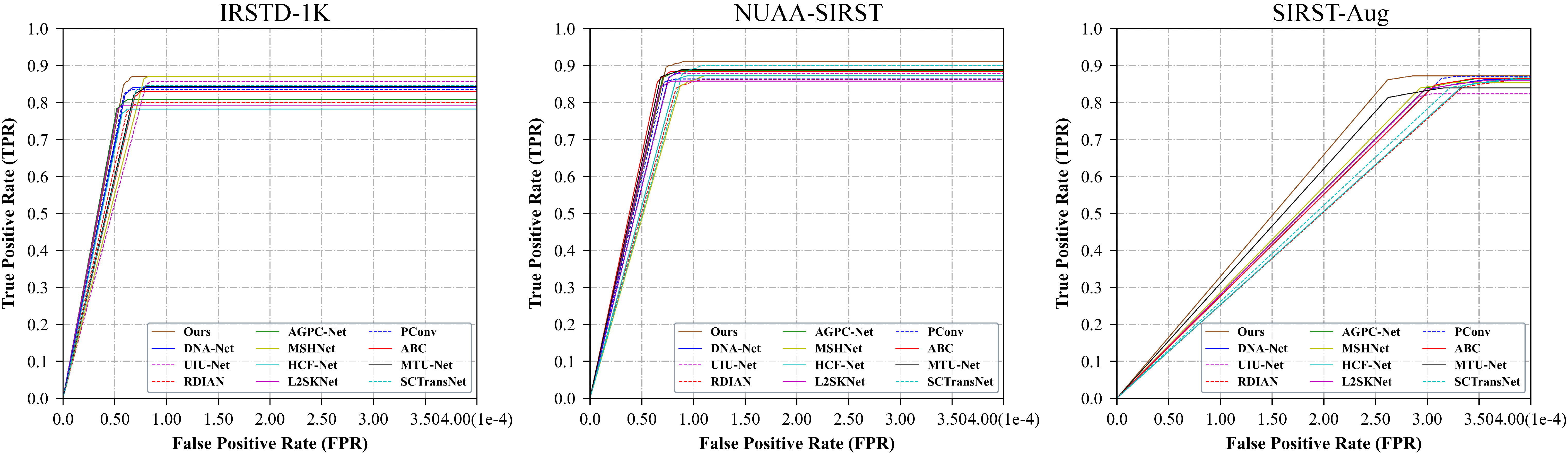}
	\caption{ROC curves comparison across three datasets.} 
	\label{Fig.5}
\end{figure}
Comparative evaluations on three datasets validate our method's effectiveness. Quantitative results in Table~\ref{tab:1} show that MDAFNet achieves superior performance on all datasets, confirming our architecture's effectiveness and robustness. Fig.~\ref{Fig.5} presents Receiver operating characteristic (ROC) curves across three benchmarks, where Ours consistently attains superior true positive rates at reduced false positive rate levels, further demonstrating enhanced detection reliability.

\subsubsection{Qualitative Evaluations Results}

Fig.~\ref{Fig.4} presents visualization of two representative examples. Under complex background interference and small target scenarios, existing deep learning methods typically exhibit high false alarm rates and miss detections. Conversely, MDAFNet accurately identifies all genuine targets with more complete boundaries, effectively suppresses background noise, and demonstrates superior detection precision and robustness.
\begin{table*}[!t]
	\centering
	\caption{Quantitative results compared against advanced approaches on IRSTD-1K, NUAA-SIRST, and SIRST-Aug. \textit{IoU}~(\%), \textit{Pd}~(\%), and \textit{Fa}~($\times 10^{-6}$) are reported, with \textbf{\textcolor{red}{red}} marking the best.}
	\label{tab:1}
	\renewcommand{\arraystretch}{1}
	\small
	\setlength{\tabcolsep}{12pt}
	\resizebox{0.98\textwidth}{!}{
	\begin{tabular}{@{}l@{\hspace{8pt}}c@{\hspace{6pt}}*{9}{c}@{}}
		\toprule
		\multirow{2.5}{*}{\textbf{Method}} & \multirow{2.5}{*}{\textbf{Publication}}
		& \multicolumn{3}{c}{\textbf{IRSTD-1K}} 
		& \multicolumn{3}{c}{\textbf{NUAA-SIRST}} 
		& \multicolumn{3}{c@{}}{\textbf{SIRST-Aug}} \\
		\cmidrule(lr){3-5} \cmidrule(lr){6-8} \cmidrule(l){9-11}
		& & IoU$\uparrow$ & P$_d$$\uparrow$ & F$_a$$\downarrow$ 
		& IoU$\uparrow$ & P$_d$$\uparrow$ & F$_a$$\downarrow$
		& IoU$\uparrow$ & P$_d$$\uparrow$ & F$_a$$\downarrow$ \\
		\midrule
		\rowcolor{blue!5}
		\multicolumn{11}{@{}l}{\textit{\textbf{CNN-based Approaches}}} \\
		\midrule
		DNA-Net~\cite{9864119}  & TIP'22 & 67.54 & 92.18 & 11.77 & 77.04 & 99.08 & 20.05 & 72.27 & \textbf{\textcolor{blue}{98.35}} & 51.85 \\
		UIU-Net~\cite{9989433} & TIP'22 & 65.59 & 88.44 & 14.73 & 76.83 & 98.17 & 11.71 & 70.64 & 96.56 & 63.47 \\
		RDIAN~\cite{10011452} & TGRS'23 & 63.40 & 92.86 & 11.54 & 74.08 & \textbf{\textcolor{blue}{100.00}} & 19.87 & 71.45 & 98.07 & 116.58 \\
		AGPC-Net~\cite{10024907} & TAES,'23 & 65.93 & 91.15 & \textbf{\textcolor{blue}{11.32}} & \textbf{\textcolor{blue}{77.47}} & \textbf{\textcolor{blue}{100.00}} & \textbf{\textcolor{blue}{3.02}} & 72.73 & 95.46 & 128.73 \\
		MSHNet~\cite{10658560} & CVPR'24 & 67.16 & \textbf{\textcolor{blue}{93.88}} & 15.03 & 73.65 & 99.08 & 19.16 & 72.64 & 96.56 & 91.55 \\
		HCF-Net~\cite{10687431} & ICME'24 & 63.72 & 89.46 & 17.38 & 75.43 & 98.17 & 11.18 & 71.69 & 97.39 & \textbf{\textcolor{blue}{37.77}} \\
		L2SKNet~\cite{10813615} & TGRS'25 & 65.67 & 92.18 & 23.99 & 75.52 & 98.17 & 17.21 & 73.26 & 97.94 & 50.34 \\
		PConv~\cite{yang2025pinwheel} & AAAI'25 & \textbf{\textcolor{blue}{67.58}} & \textbf{\textcolor{blue}{93.88}} & 12.22 & 75.98 & 98.17 & 4.26 & \textbf{\textcolor{blue}{74.04}} & 93.95 & 288.71 \\
		\midrule
		\rowcolor{teal!10}
		\multicolumn{11}{@{}l}{\textit{\textbf{Hybrid Approaches}}} \\
		\midrule
		ABC~\cite{10219645} &  ICME'23 & 65.13 & 88.43 & \textbf{\textcolor{teal}{14.96}} & \textbf{\textcolor{teal}{77.59}} & 98.17 & 6.74 & 72.86 & \textbf{\textcolor{teal}{98.76}} & 56.64 \\
		MTU-Net~\cite{10011449} & TGRS'23 & 65.44 & 87.75 & 16.40 & 77.46 & \textbf{\textcolor{teal}{100.00}} & \textbf{\textcolor{teal}{5.68}} & 71.59 & 96.42 & \textbf{\textcolor{teal}{37.57}} \\
		SCTransNet~\cite{10486932} & TGRS'24 & \textbf{\textcolor{teal}{66.07}} & \textbf{\textcolor{teal}{92.52}} & 15.79 & 76.18 & 99.08 & 26.97 & \textbf{\textcolor{teal}{73.79}} & 98.21 & 46.67 \\
		\midrule
		\textbf{MDAFNet (Ours)} & \textbf{--} &  \textbf{\textcolor{red}{70.11}} & \textbf{\textcolor{red}{95.92}} & \textbf{\textcolor{red}{8.43}} 
		& \textbf{\textcolor{red}{79.42}} & \textbf{\textcolor{red}{100.00}} & \textbf{\textcolor{red}{3.90}} 
		& \textbf{\textcolor{red}{75.60}} & \textbf{\textcolor{red}{99.45}} & \textbf{\textcolor{red}{15.15}} \\
		\bottomrule
	\end{tabular}%
} 
\end{table*}
\begin{figure}[t]
	\centering
	\renewcommand{\arraystretch}{0.9}
	\includegraphics[width=0.48\textwidth]{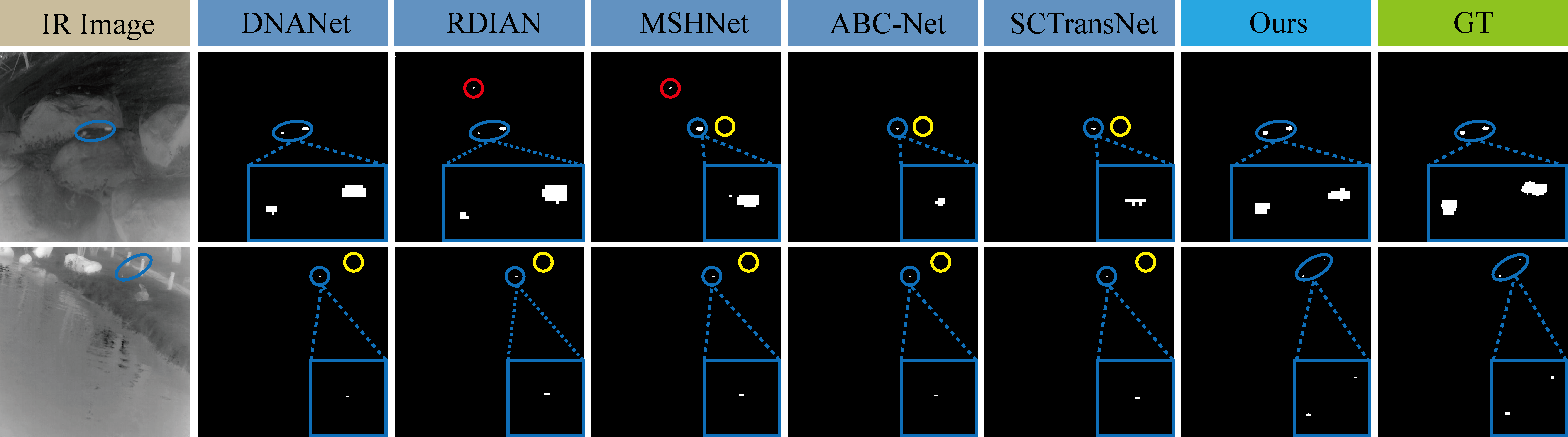}
	\caption{Visual comparison of detection results for the proposed method against other mainstream approaches. Correctly detected targets are marked in blue, missed targets in yellow, and false alarms in red.} 
	\label{Fig.4}
\end{figure}
\subsection{Ablation Study}
\subsubsection{Conduct ablation on major components}
Comprehensive ablation studies validate each module's contribution. As shown in Table~\ref{tab:2}, individually adding either the MSDE or DAFE module to the baseline brings performance improvements, while the complete MDAFNet integrating both modules achieves optimal performance, significantly outperforming any single-module configuration, fully validating the effectiveness and necessity of the MSDE and DAFE modules.

\begin{table}[h]
	\renewcommand{\arraystretch}{0.9}
	\centering
	\caption{Ablation study on IRSTD-1K with \textit{IoU}~(\%), \textit{Pd}~(\%), and \textit{Fa}~($\times 10^{-6}$), Para. (M), and FLOPs (G).}
	\label{tab:2}
	\resizebox{0.48\textwidth}{!}{%
		\begin{tabular}{l|ccc|cc}
			\toprule
			\textbf{Module} & \textbf{IoU} & \textbf{P$_d$} & \textbf{F$_a$} &\textbf{Para.} &\textbf{FLOPs}\\
			\midrule
			Baseline  & 67.16 & 93.88 & 15.03 & 4.07 & 6.11 \\
			Baseline+MSDE  & 68.26 & 94.90 & 11.99 & 4.19 & 6.59 \\
			Baseline+DAFE  & 68.86 & 94.56 &  9.41& 4.33 & 6.36\\
		    Baseline+MSDE+DAFE  & \textbf{\textcolor{red}{70.11}} & \textbf{\textcolor{red}{95.92}} & \textbf{\textcolor{red}{8.43}}  &4.45 & 6.83 \\
			\bottomrule
		\end{tabular}%
	}
\end{table}

\subsubsection{Selection of multi-scale branch numbers of MSDE}
To examine how multi-scale branch quantity in MSDE affects performance, comparative evaluations are conducted using the IRSTD-1K dataset. Table~\ref{tab:3} demonstrates that optimal results are obtained with Width=4. Too few differential edge branches cannot sufficiently extract edge information, while too many branches lead to feature redundancy. Balancing detection accuracy and computational cost, the width parameter of MSDE is set to 4.
\begin{table}[!t]
	\renewcommand{\arraystretch}{0.9}
	\centering
	\caption{Ablation study on the width parameter of MSDE.}
	\label{tab:3}
	\setlength{\tabcolsep}{8pt}
	\resizebox{0.48\textwidth}{!}{%
	\begin{tabular}{c|ccc|cc}
		\toprule
		\textbf{Width} & \textbf{IoU} & \textbf{P$_d$} & \textbf{F$_a$}  &\textbf{Para.} &\textbf{FLOPs}\\
		\midrule
		3 & 67.85 & 93.54 & 11.39 & 4.187 & 6.565 \\
		4 & \textbf{\textcolor{red}{68.26}} & \textbf{\textcolor{red}{94.90}} & 11.99 & 4.188 & 6.587 \\
		5 & 67.60 & 93.88 & 15.03 & 4.188 & 6.609 \\
		6 & 67.94 & 91.84 & \textbf{\textcolor{red}{10.93}} & 4.188 & 6.631 \\
		\bottomrule
	\end{tabular}	}
\end{table}
\subsubsection{Triple-path fusion effectiveness validation}
Comparative experiments involving four distinct fusion approaches are conducted to evaluate our triple-path strategy. As shown in Table~\ref{tab:4}, basic fusion methods such as simple addition and element-wise multiplication fail to effectively improve performance. In contrast, the proposed adaptive triple-path fusion strategy achieves optimal performance by integrating features through three adaptive pathways, effectively compensating for edge information loss.
\begin{table}[!t]
	\renewcommand{\arraystretch}{0.9}
	\centering
	\caption{Triple-path fusion effectiveness validation.}
	\label{tab:4}
	\setlength{\tabcolsep}{6pt}
	\resizebox{0.48\textwidth}{!}{%
	\begin{tabular}{c|l|ccc}
		\toprule
		\textbf{No.} & \textbf{Module} & \textbf{IoU} & \textbf{P$_d$} & \textbf{F$_a$} \\
		\midrule
		1 & None & 67.16 & 93.88 & 15.03 \\
		2 & Simple addition  & 66.97 & 92.18 & 14.27 \\
		3 & Element-wise multiply  & 67.20 & 93.54 & 16.40 \\
		4 & Adaptive Dual-path  & 67.17 & 93.20 & 14.20 \\
		5 & \textbf{Adaptive Triple-path (Ours)} & \textbf{\textcolor{red}{68.26}} & \textbf{\textcolor{red}{94.90}} & \textbf{\textcolor{red}{11.99}} \\
		\bottomrule
	\end{tabular}}
\end{table}
\subsubsection{Ablation study on removing individual components from DAFE}
Ablation experiments evaluate DAFE's component contributions by sequentially removing each module. As shown in Table~\ref{tab:5}, removing any component leads to performance degradation. Specifically, removing DWT/IDWT significantly reduces detection rate, removing MSKP increases $F_a$, and removing AFM causes the most significant performance drop, validating its critical role in suppressing high-frequency noise. Each DAFE component positively impacts overall detection performance.

\begin{table}[!t]
	\renewcommand{\arraystretch}{0.90}
	\centering
	\caption{Ablation study of DAFE components.}
	\label{tab:5}
	\resizebox{0.48\textwidth}{!}{%
	\begin{tabular}{l|ccc|cc}
		\toprule
		\textbf{Module} & \textbf{IoU} & \textbf{P$_d$} & \textbf{F$_a$}  &\textbf{Para.} &\textbf{FLOPs}\\
		\midrule
		\textbf{DAFE} & \textbf{\textcolor{red}{68.86}} & \textbf{\textcolor{red}{94.56}} & \textbf{\textcolor{red}{9.41}} & 4.331 & 6.357\\
		w/o DWT/IDWT & 67.81 & 91.50 & 10.02 &4.153 & 6.504 \\
		w/o MSKP & 67.28 & \textbf{\textcolor{red}{94.56}} & 12.98 & 4.440   & 6.413\\
		w/o AFM & 67.64 & 93.20 & 14.42 & 4.331  & 6.353 \\
		w/o Adaptive Residual & 68.69 & 94.22 & 9.717 & 4.331 & 6.357 \\
		\bottomrule
	\end{tabular}}
\end{table}

\section{Conclusion}
We present MDAFNet for IRSTD. The MSDE module effectively preserves target geometric detail integrity through a multi-scale differential edge enhancement mechanism. Additionally, the DAFE module achieves effective separation of high-frequency targets and noise through adaptive frequency guidance. Experiments conducted on several benchmark datasets reveal that MDAFNet achieves satisfactory results, confirming our approach's effectiveness. Future work intends to explore lightweight network architectures to adapt to real-time detection scenarios.

\bibliographystyle{IEEEtran}
\bibliography{reference} 

\end{document}